\documentclass[letterpaper, 10 pt, conference]{ieeeconf}  
\usepackage{FG2021}
\usepackage{float}
\usepackage{graphicx}
\usepackage{hyperref}
\usepackage{nopageno}
\FGfinalcopy 

\IEEEoverridecommandlockouts                              
\title{\LARGE \bf
The Many Faces of Anger: A Multicultural Video Dataset of Negative Emotions in the Wild (MFA-Wild)
}

\author{\parbox{16cm}{\centering
    {\large Roya Javadi$^1$ and Angelica Lim$^1$}\\
    {\normalsize
    $^1$School of Computing Science, Simon Fraser University, Burnaby, Canada}}
    \thanks{We acknowledge the support of the Natural Sciences and Engineering Research Council of Canada (NSERC), RGPIN/06908-2019. We also thank people who helped us in collection and labeling of the data, especially Emma Hughson.}
}

\begin{document}


\ifFGfinal
\thispagestyle{empty}
\pagestyle{empty}
\else
\author{Anonymous FG2021 submission\\ Paper ID \FGPaperID \\}
\pagestyle{plain}
\fi
\maketitle

\begin{abstract}

The portrayal of negative emotions such as anger can vary widely between cultures and contexts, depending on the acceptability of expressing full-blown emotions rather than suppression to maintain harmony. The majority of emotional datasets collect data under the broad label ``anger", but social signals can range from annoyed, contemptuous, angry, furious, hateful, and more. In this work, we curated the first in-the-wild multicultural video dataset of emotions, and deeply explored anger-related emotional expressions by asking culture-fluent annotators to label the videos with 6 labels and 13 emojis in a multi-label framework. We provide a baseline multi-label classifier on our dataset, and show how emojis can be effectively used as a language-agnostic tool for annotation.


\end{abstract}

\section{Introduction}

Anger is a basic emotion suggested to be found around the world. Despite the universal theories that associated prototypical expressions such as bared teeth to anger, depictions of anger come in a variety of social signals and arousal levels \cite{cite1}.  Furthermore, some research \cite{cite2, cite3, cite4} suggests that more collectivist cultures tend to suppress negative emotions to maintain harmony. These reasons imply that correctly recognizing anger in videos is highly dependent on the samples fed into models for emotion recognition tasks. 

Automatic emotion expression recognition for anger samples is relatively low, and deserves more investigation. In \cite{siqueira2020efficient} for example, authors found that using both AffectNet and FER+, resulted in accuracy between 45\% and 54\% for contempt, anger, and disgust on the AffectNet dataset, while happiness had 77\% accuracy rate. Similarly in \cite{Kosti_2019} where images were multi-labeled with 26 categories, after `aversion' with an average precision of 7.48, anger was the worst recognized basic emotion with an average precision of 9.49, followed by fear (14.14), surprise (18.81), sadness (19.66) and happiness (60.69). Hence, a deeper understanding of the anger emotion is necessary.

Most of the work in this field is on images, but the dynamics of face and head are also important. Therefore, we decide to focus on dynamic representations of emotions, such as videos. A major challenge in emotion recognition in videos is to find a high-quality dataset, which includes both high-quality videos and their corresponding labels. In particular, it is important to gather in-the-wild datasets, as lab-acted datasets such as  \cite{amsterdam-dataset} may not well represent the variety of possible expressions. Among in-the-wild datasets, GIFGIF+ \cite{gifgif} introduced by MIT is a relatively massive and realistic video dataset that consists of more than 23,000 animated GIFs over 17 emotions. The authors' dataset collection method is semi-automatic, meaning that they used both human labour and clustering techniques to label GIF samples. Aff-Wild2 \cite{kollias2018affwild2} is another in-the-wild video dataset containing 260 videos that are used for estimation of the valence and arousal emotion dimensions. A variety of subjects, movements, and context is also another key point in video datasets. The Affectiva-MIT Facial Expression Dataset (AM-FED) database \cite{AM-FED} contains 242 facial videos of people watching Super Bowl commercials using their webcam. It has frame-by-frame annotations of 14 Action Units (AU), head movements, and facial landmarks, but there is not much variance in head poses and subjects, as all participants are reacting to videos while sitting in front of a computer. 

Psychological research indicates that people express emotions differently depending on their age, gender or culture \cite{cite31}. A few datasets have focused on underrepresented groups in emotion recognition, namely EmoReact \cite{emoreact}, which contains videos of children in the wild reacting to objects and answering questions about them, and ElderReact  \cite{elderreact} where older adults react to videos and express their opinion. Both of them collected in-the-wild videos from the same channel on YouTube, containing videos of people seated at a desk.

Annotation schemes in affective computing still have room for improvement. For instance, the broad emotion label of anger is problematic for several reasons.  Firstly, there is a range of anger-related phenomena which could be more precisely recognized with  a more specific word. Similar to how the word ``mammal" can encompass a wide range of animals, more specific words such as ``cat" or ``dog" can provide better labels. Here, we consider labels such as annoyed, furious, hatred, contempt and disgust. 
Secondly,  emotion ``readings" can be highly subjective theory of mind exercises, which can depend on the annotator or their own internal state. An alternate scheme is to provide objective behavioral descriptions of facial expressions (e.g. frowning, rolling eyes), similar to AU recognition but at a higher level. Here, we explore representations of social signals using emojis, which in addition are language-agnostic.
Using emojis for labeling has gained popularity recently. Saheb Jam et al. \cite{Saheb_Jam_2021} used emojis to annotate emotional expressions in videos of interaction between a human and a robot.  Vemulapalli et al. have used emojis to build an embedding space for facial expressions \cite{emoji-emb}. Herein, we used 13 emojis related to 6 emotion labels of ‘annoyed’, ‘anger’, ‘fury’, ‘hatred’, ‘disgust’, ‘contempt’ in order to investigate whether mapping emojis to their emotion category can improve language-agnostic annotations. Thirdly, emotional expressions can be mixed \cite{DuE1454}. While fields such as text, speech, and music emotion processing have readily accepted the multi-label paradigm, video-based approaches still assume one label per sample, throwing out data that does not have high inter-rater agreement or use single-label classifiers on datasets that have multi-labels for each sample. As noted by \cite{ml_speech}, ``ambiguous, subtle expressions of emotion, which often obtain no majority agreement from human annotators, are prevalent in the real world". Since we are working on in-the-wild data, a multi-label baseline is an important component in this work.

To summarize, the main contributions of this study are collecting data from underrepresented cultures in the anger category, as well as annotating them by people from the same culture. The majority of datasets usually focus on Western-Caucasian subjects and we rarely see other cultures or ethnicities such as East-Asian or West-Asian. In this research, we tried to bridge this gap by collecting data from underrepresented cultures from Middle East, which  to the best of our knowledge has never been done before. A few works such as \cite{Benitez-Garcia} designed feature extractors and a classifier on a multi-cultural (East-Asian and Western-Caucasian) image dataset.  Khanh et al. also collected a Korean emotion dataset from Korean movies and used a Multi-Layer Perceptron to classify them into basic emotions. They showed that training the model using Korean videos and testing on English videos and vice-versa yielded the worst result \cite{korean-dataset}. 
Another contribution of ours is a collection of social signals for each video in the dataset, aiming to identify culture-dependent facial expressions in emotions.

In this paper, we first describe the process of raw data collection. Then, we outline the emotion and emoji annotation procedure. Finally, we present dataset statistics and a baseline classification for future benchmarking.

\section{Data collection}

The massive amount of videos on YouTube is a great asset for researchers. A major challenge that remains to be addressed is extending videos with in-the-wild or close to in-the-wild emotions.  
We collected more than 200 videos on YouTube from Persian and North American (NA) cultures. All Persian videos were collected from TV series and movies accessible on YouTube, and NA videos included movies, ``vlog" style content, reality television shows (e.g., Dance Moms), and talk shows (e.g., The Late Show with James Corden). Since some videos were too long for our purpose or had multiple actors, we trimmed,  split, or removed them. In the end, there were 97 videos for Persian and 104 videos for NA culture, each lasting between 1 to 10 seconds.

Our main  challenge in label collection was that crowd-sourcing platforms such as Amazon Mechanical Turk (AMT) do not provide a facility to choose the culture of annotators. Thus, we decided to design and implement a web interface and collect culturally fluent annotators in Canada. Annotators were 19+ years old residents of Canada who were fluent in English, as well as Farsi if annotating Persian videos.

We designed an interface using Flask\footnote{https://flask.palletsprojects.com/} and an SQLite database. Each annotator registered on the website and accepted a Research Consent form. We only asked for general information about the culture, language, and perceived individualism of people. No directly identifiable information was collected during the study. The participants were allowed to withdraw from the study at any time. An Amazon gift card code corresponding to approximately the minimum wage in Canada was emitted upon completion of or withdrawal from the study, and this study was approved by the university research ethics board. 
In total, we recruited 10 people from each culture via social media. The audio was removed from the videos. Each user annotated half of the videos of their culture, resulting in 5 annotations per clip.

\subsubsection{Emotion Labels}
We defined 6 negative emotions for labels: Contemptuous, Annoyed, Anger, Hatred, Furious, and Disgusted. We also added a “None” option. Annotators were allowed to select more than one label and leave their idea or thoughts in a comment box. The annotations were done independently and blindly, meaning that annotators did not have any information about each other's labels.

\begin{figure}[t]
\centering
\includegraphics[scale=0.6]{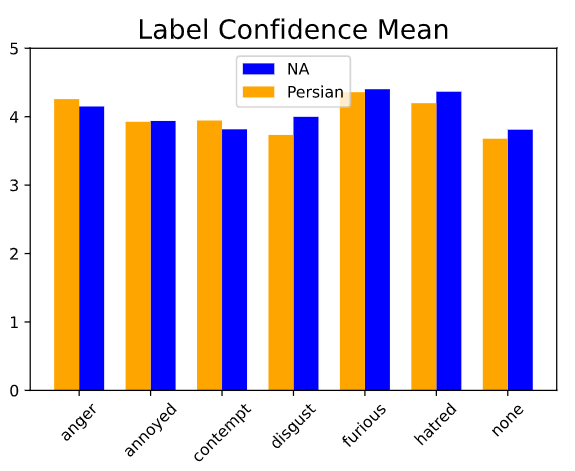}
\caption{Annotators' confidence for labels}

\label{fig:confidence}
\end{figure}

\begin{figure}[t]
\centering
\includegraphics[scale=0.4]{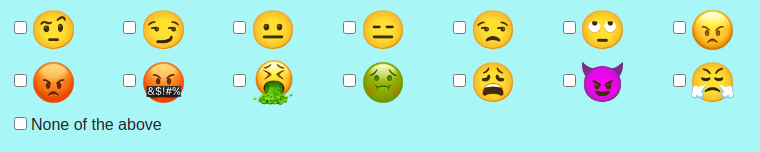}
\caption{Emoji labels}
\label{fig:emojis}
\end{figure}

\subsubsection{Emoji Labels}
\begin{figure}[t]
 \centering
\includegraphics[scale=0.6]{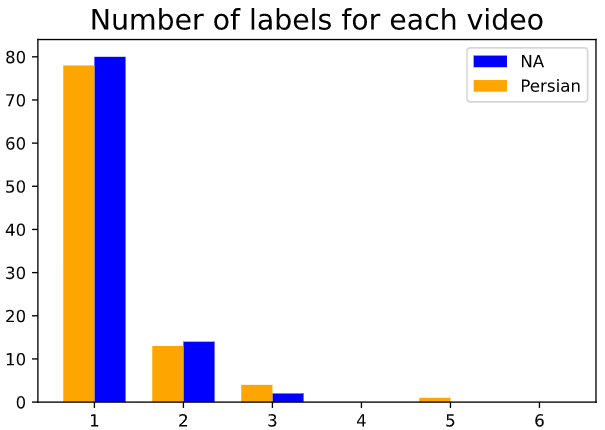}
\caption{Distribution of the number of emotion labels}
\label{fig:labels-dist}
\end{figure}

 In this study, we also included 14 emoji annotations to extract underlying subtle social signals of each emotion. The emojis that users could select is presented in Fig. \ref{fig:emojis}. Pilot annotations among researchers suggested difficulty in obtaining unanimity in labeling; in many samples, people selected multiple emotions. This gave us a clue that allowing multiple labels for videos would be better than restricting the labels to only one. We accumulated all the emotion and emoji labels belonging to each video and did majority voting to derive final labels. If two or more labels had equal votes, we attached all those labels to the video. For example, a video that has 2 votes for annoyed, 2 votes for anger, and 1 vote for contempt, is assigned both anger and annoyed. Fig. \ref{fig:labels-dist} shows the distribution of the number of labels.

 \subsubsection{Social Signals Annotation}
 In order to explore what social signals can be found in emotional expressions, one researcher from NA cultural background and one from Persian culture extracted visible face and body expressions for each video. Since social signal annotations such as raised brows, arms crossed, turning head, etc. are relatively objective (compared to emotion labels), both researchers reviewed all videos for this annotation step. Overall, they had a consensus for over 90\% of videos. The analysis we conducted on these social signals aimed to identify the cultural differences in emotion expression.
 
 \section{Dataset Statistics}
\begin{figure}[t]
 \centering
\includegraphics[scale=0.6]{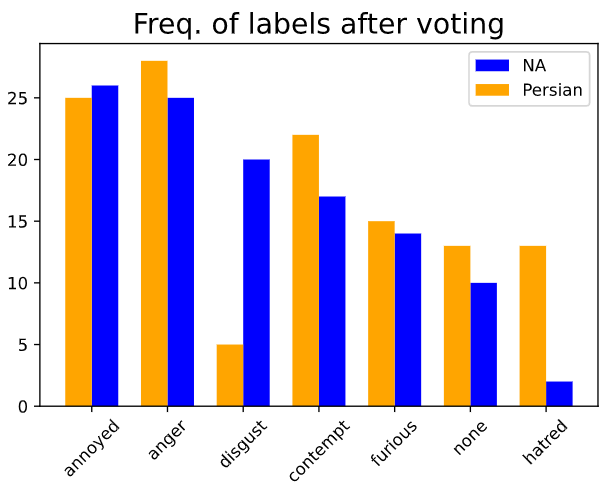}
\caption{Frequency of emotion labels}
\label{fig:emotion-freq}
\end{figure}

\begin{figure*}
\centering
\includegraphics[scale=0.5]{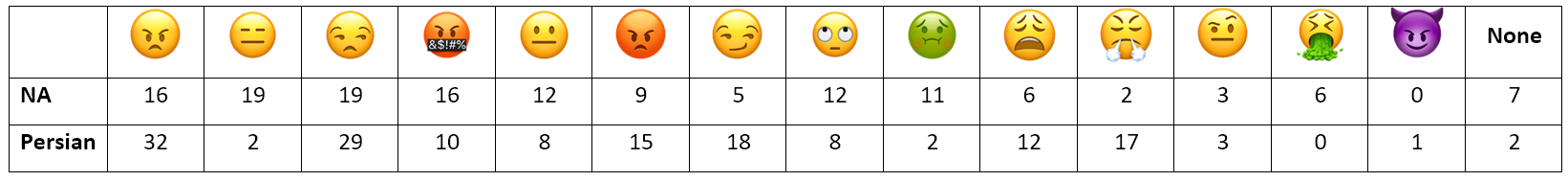}
\caption{Frequency of emoji labels}
\label{fig:emoji-freq}
\end{figure*}

  Initially, our dataset aimed to cover the emotions of Contempt, Anger, and Disgust (also known as the CAD triad \cite{rozin-cad}). The first steps of annotation revealed that annotators perceived some fine-grained emotions like annoyed, as well as complex emotions such as disgust-anger or contempt-disgust. Therefore, we decided to further refine the labels and allow people to choose several emotions. We emphasize that the purpose of this study is not to obtain perfect agreement, but to consider that, similar to \cite{ml_speech}, the distribution over annotators \emph{is} the ground truth for this challenging in-the-wild data. Nevertheless, as a descriptive measure, we used Krippendorff’s Alpha (used when multiple labels can be chosen) to calculate the agreement between annotations. The agreement scores for NA and Persian emotion dataset were 0.252 and 0.076, respectively. Since Jeni et al. \cite{jeni2013} indicated that label imbalance can dramatically affect metrics such as Krippendorff’s  Alpha, and we can see that labels were indeed imbalanced when observing the number of videos for each emotion/emoji in Fig. \ref{fig:emotion-freq} and Fig. \ref{fig:emoji-freq}.

\begin{figure}[t]
\centering
\includegraphics[scale=0.6]{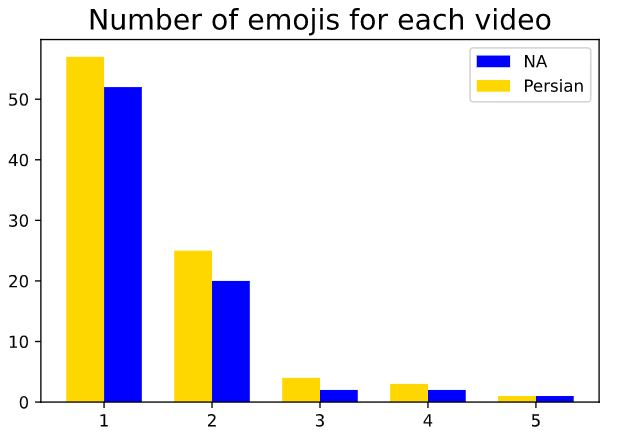}
\caption{Distribution of the number of emoji labels}
\label{fig:emoji-dist}
\end{figure}

We calculated the mean of the annotator’s confidence ratings for each label, which you can see in Fig. \ref{fig:confidence}. The noticeable difference between the two cultures is the confidence for disgust, which is higher in NA. That is because most disgust videos in NA dataset were reactions to foods, whereas in Persian ‘disgust’ videos were social disgust, hence more challenging to label.

\subsection{Co-Occurrence of Emoji-Emotions}
We calculated which emojis co-occur with which emotions. Fig. \ref{fig:common-emojis} shows the emojis that appeared in more than 15\% of the videos for a specific label. Interestingly, annotators associated red emojis with higher arousal forms of anger, which supports the research that red faces map to higher levels of anger \cite{ikeda2020influence}. Also, annotators chose other emojis like \includegraphics[scale=0.18]{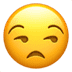} over \includegraphics[scale=0.18]{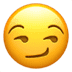} for contempt. In NA dataset, we see an abundance of ‘nauseated’ emoji for disgust while in Persian annotations there is none; this is due to the videos in NA that react to food, as we mentioned.
Another interesting finding from this analysis is that the label ``contempt" is not fine-grained enough; users selected the \includegraphics[scale=0.18]{images/unamused.png}emoji when it was combined with social disgust, while they preferred \includegraphics[scale=0.18]{images/smirk.png} for contempt alone.

\begin{figure}[t]
\centering
\includegraphics[scale=0.4]{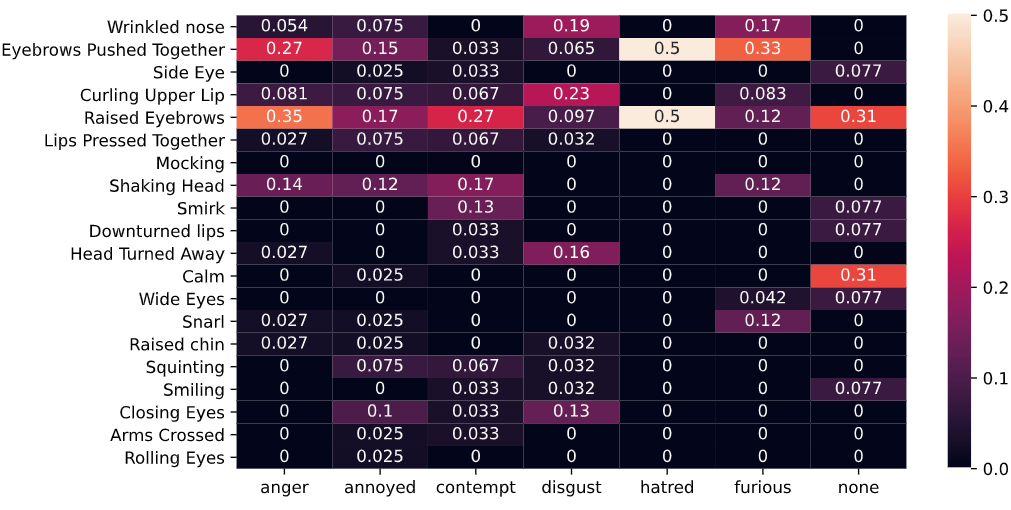}
\caption{Co-occurrences of social signals for each emotion in NA videos}
\label{fig:ss_emotion_na}
\end{figure}

\begin{figure}[t]
\centering
\includegraphics[scale=0.4]{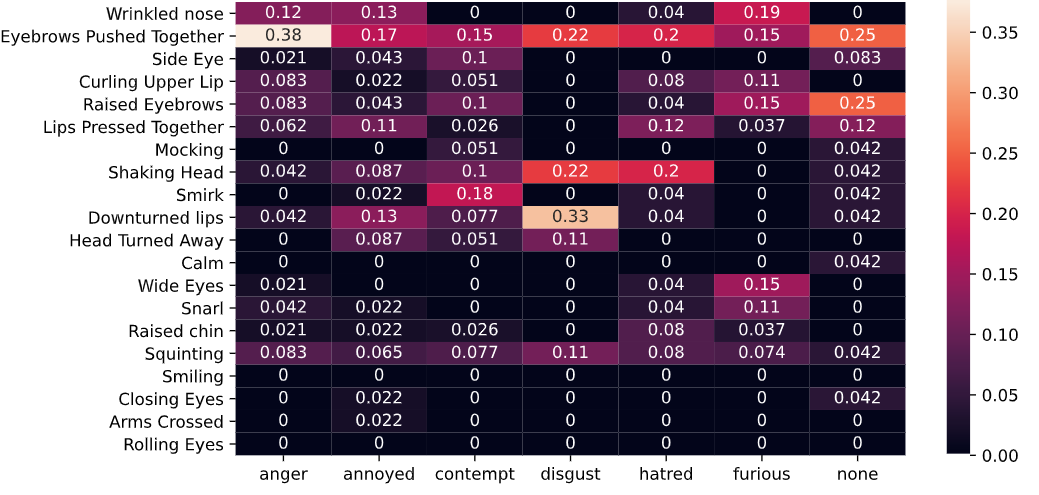}
\caption{Co-occurrences of social signals for each emotion in Persian videos}
\label{fig:ss_emotion_persian}
\end{figure}

\subsection{Correlation between emotions}
\begin{figure}[t]
\centering
\includegraphics[scale=0.4]{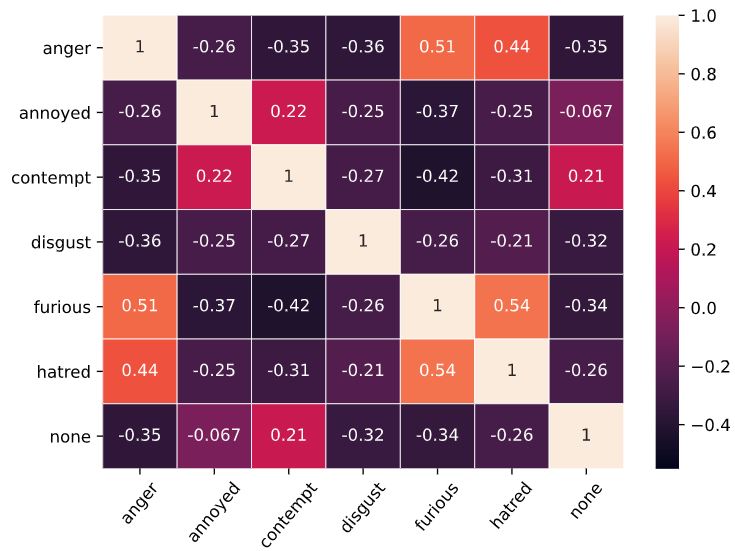}
\caption{Correlation between annotators' assigned labels - NA}
\label{fig:na-corr}
\end{figure}

\begin{figure}[t]
\centering
\includegraphics[scale=0.4]{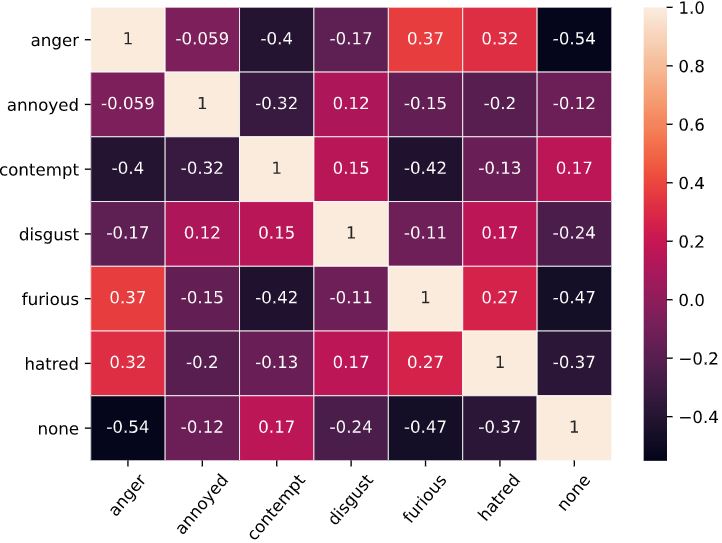}
\caption{Correlation between annotators' assigned labels - Persian}
\label{fig:p-corr}
\end{figure}

We think the correlation between emotions may shed light on how emotions are related to each other, so we computed the Pearson correlation between the labels given by annotators (i.e. before voting the labels). The results are in Fig. \ref{fig:na-corr} and Fig. \ref{fig:p-corr}, which provides us another view of the relationship between negative emotions. It is not surprising that anger, furious and hatred have a relatively high positive correlation. It is worth noting that disgust and hatred have a positive correlation in the Persian dataset while their correlation is negative in NA dataset. Disgust and contempt show different patterns as well. The correlation between them is positive in Persian and negative in NA. Such differences may suggest that we should distinguish between ‘social disgust’ and ‘physical disgust’. Social disgust might be a mix of hatred, annoyance and contempt according to the result, whereas ‘physical disgust’ can be considered a basic emotion \cite{ekman1992argument}. These tables also show the importance of having multiple labels for samples in order to capture compound negative feelings such as 'angry and disgusted' or 'contemptuous and disgusted', which are seen in Persian dataset labels.

\subsection{Action Unit Analysis}
For all videos under each emotion category, we calculated the mean of peaks of each AU (one value per AU, per video) and visualized it using a radar chart in Table \ref{tab:radar}. This table can give a better view of differences between cultures. In our dataset, the AU values for Persian are generally smaller than NA. We see noticeable differences in the activated AUs in contempt, annoyed, hatred, and anger.

\begin{figure*}[t]
\centering
\includegraphics[scale=0.6]{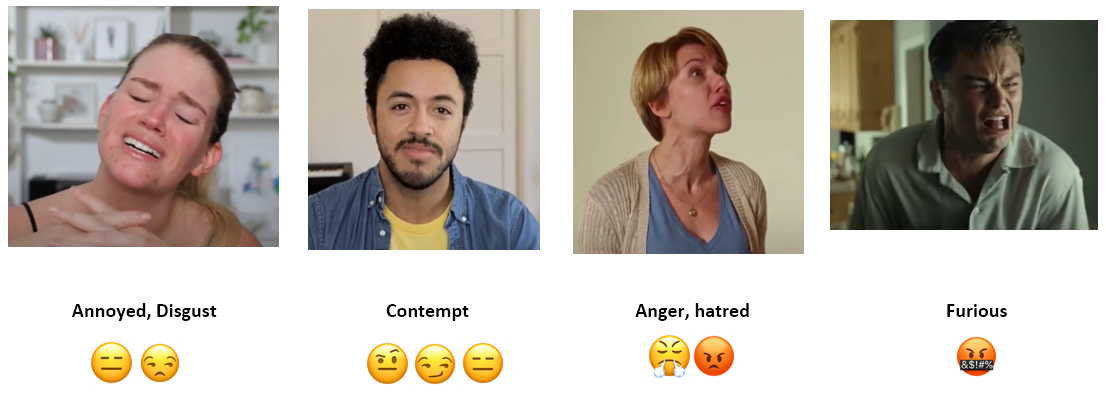}
\caption{Examples of NA dataset with their labels}
\label{fig:na_examples}
\end{figure*}

\begin{figure*}[t]
\centering
\includegraphics[scale=0.6]{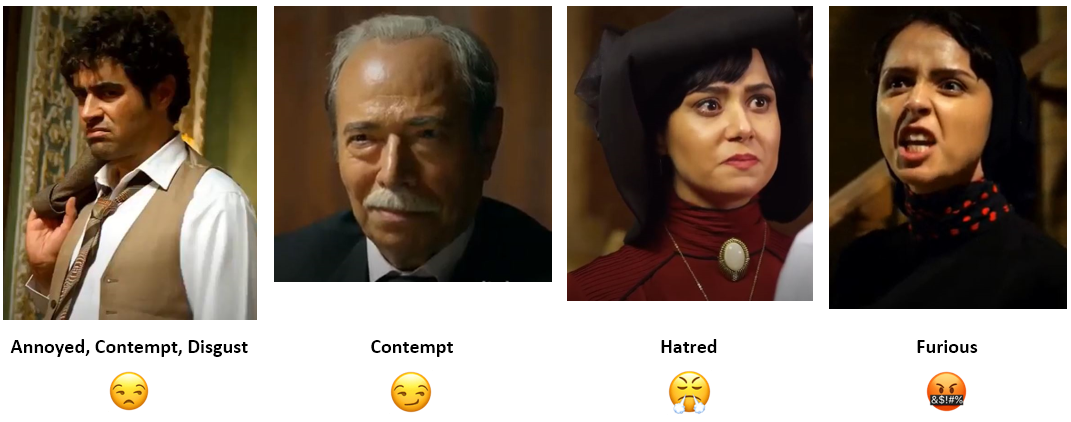}
\caption{Examples of Persian dataset with their labels}
\label{fig:p_examples}
\end{figure*}

\begin{table*}[t]
\caption{Radar charts of the mean of the peak of AUs - Numbers around each chart indicate AU index.}
  \centering
  \begin{tabular}{| c | c | c | c | c | c | c |}
  \hline
     Culture & Contempt & Annoyed & Anger & Hatred & Furious \\ [1ex] \hline
     NA & \includegraphics[scale=0.29]{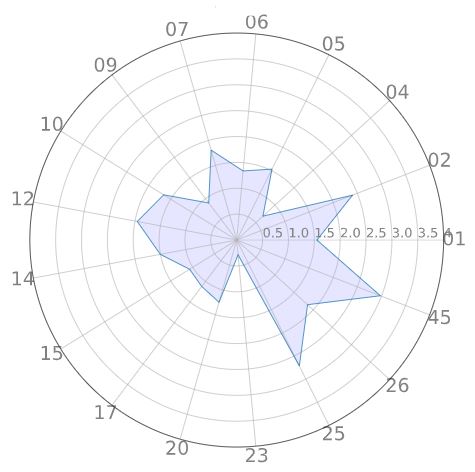}&\includegraphics[scale=0.29]{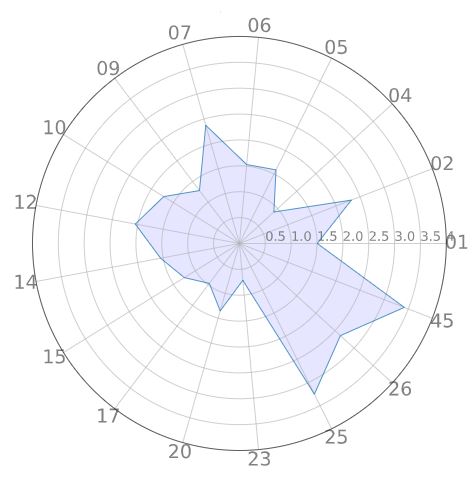} &
     \includegraphics[scale=0.28]{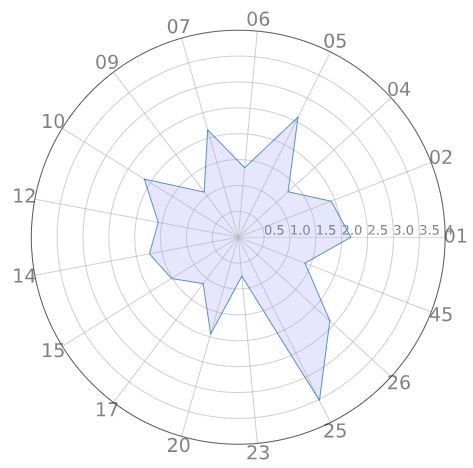} &
     \includegraphics[scale=0.28]{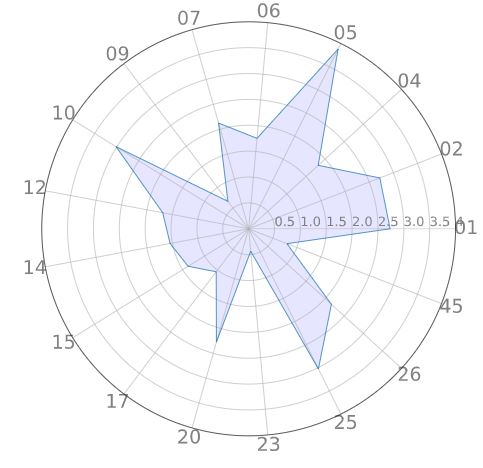} &
     \includegraphics[scale=0.28]{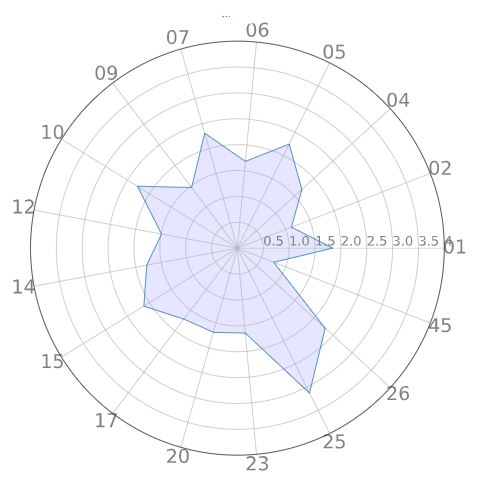} \\ [1ex] \hline
     Persian & \includegraphics[scale=0.28]{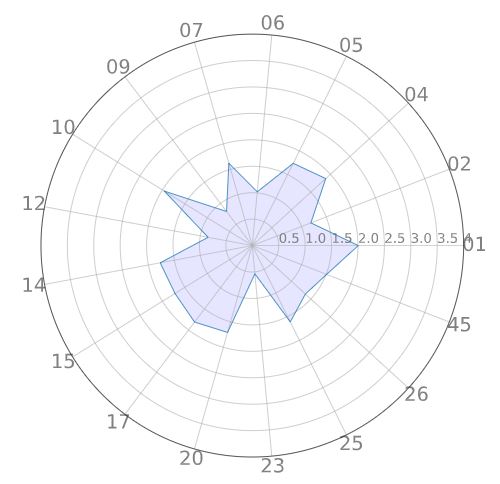}&\includegraphics[scale=0.28]{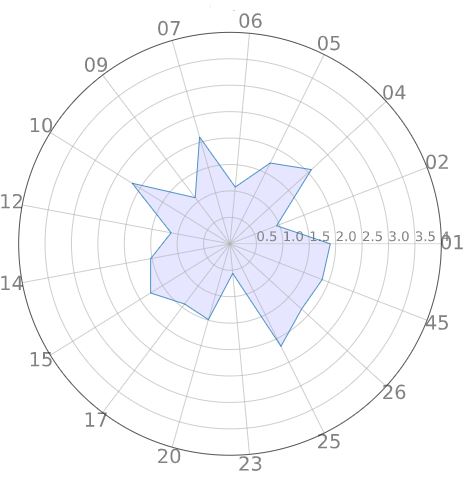} &
     \includegraphics[scale=0.28]{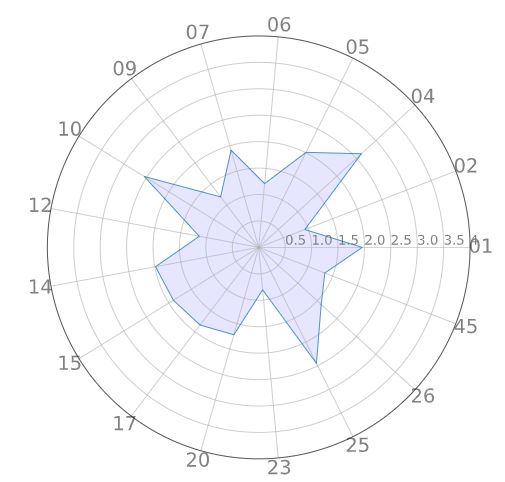} &
     \includegraphics[scale=0.28]{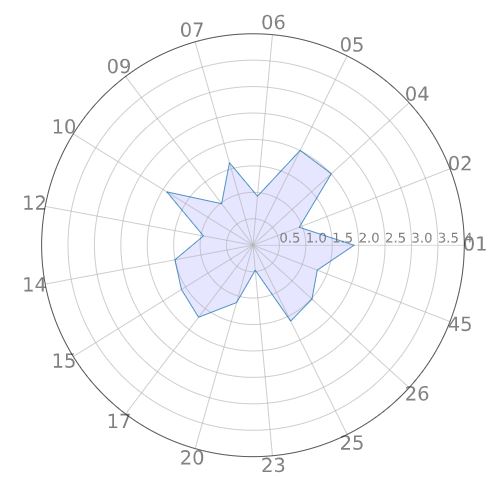} &
     \includegraphics[scale=0.28]{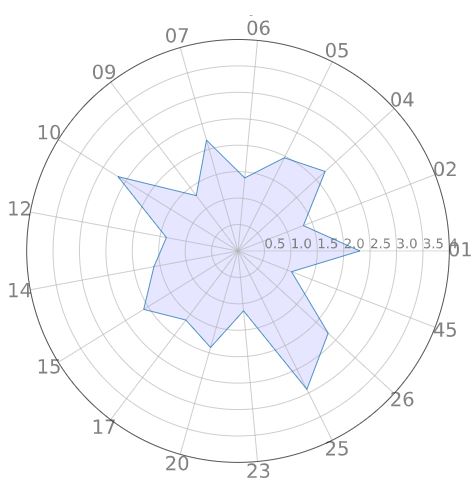} \\ [1ex] \hline
    \end{tabular}
    \label{tab:radar}
\end{table*}

\subsection{Co-Occurrence of Social Signals-Emotions}
Another purpose of this study is to identify the key social signals for each emotion. We extracted co-occurrences of facial expressions (and body movements, such as arms crossed) with final labels, presented in Fig.~\ref{fig:ss_emotion_na} for NA and Fig.~\ref{fig:ss_emotion_persian} for Persian. Each cell in the heatmap is normalized by the number of samples under each emotion class. Mocking in NA videos and Smiling social signal in Persian had 0 occurrences so their corresponding row is empty. Persian videos exhibited a wider variety of social signals except in disgust, which may be attributed to the low number of videos in this class. `Eyebrows pushed together' was the most prevalent of all social signals in all emotions, though it was more common in Persian. In the Persian results, we notice some similarities between disgust and hatred social signals.

\section{Experiments}
In this section, we describe preprocessing of the data and the baseline classification that we performed.

\subsection{Data Selection and Cleaning}
Our analysis showed that the type of disgust collected in our dataset was heterogeneous, containing both physical and social disgust, therefore, prior to doing classification, we removed the 'disgust' label, its associated emojis (\includegraphics[scale=0.18]{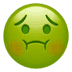} and \includegraphics[scale=0.18]{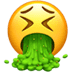}) and videos which only had 'disgust' as label, focusing on the anger affect. We also removed 2 emojis that appeared least of all: \includegraphics[scale=0.18]{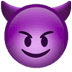} and \includegraphics[scale=0.18]{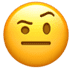}. In the end, we performed our experiments on 74 videos from NA and 91 videos from Persian culture.

Secondly, we checked the annotations for probable errors (e.g. checking the gender of actor to validate the annotation). In a few cases, the video was not properly loaded for the user, hence the actor's gender was inconsistent; we removed those annotations. In one video in which two actors appeared, some users annotated female and some male, so we removed these videos from our dataset. 

\subsection{Features}
OpenFace \cite{openface} was used to extract facial landmarks and movements, gaze direction, and head movement for each frame in a given video.  We collected 17 Action Unit (AU) attributes (AU1, AU2, AU4, AU5, AU6, AU7, AU9, AU10, AU12, AU14, AU15, AU17, AU23, AU25, AU26, AU45) describing relative values for each AU, head rotation values in 3-dimensional space and gaze direction. In addition to these features, success level, confidence level, frame number and timestamp of each frame were also collected.

During preprocessing, we only kept the frames where the face tracking resulted in a confidence level of above 85\% and a success value of 1. We performed min-max feature normalization because the range of values for the aforementioned features differed.

%
%

\begin{figure}[t]
\centering
\includegraphics[scale=0.6]{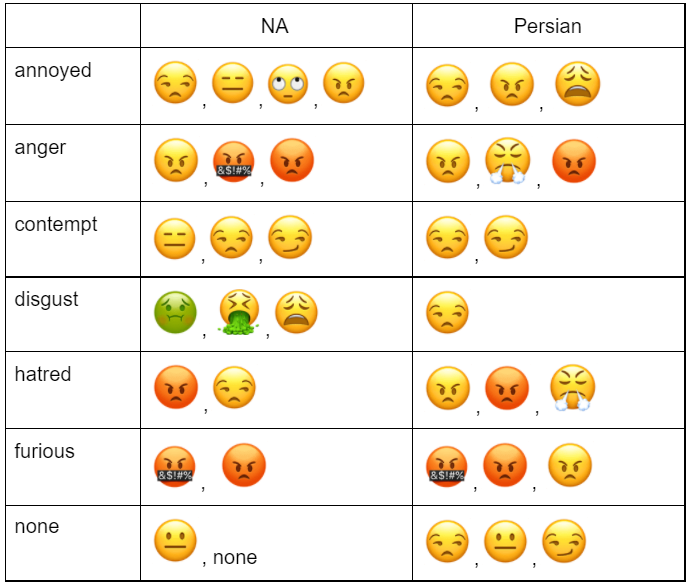}
\caption{Most common emojis for each emotion}
\label{fig:common-emojis}
\end{figure}

\begin{table}[t]
\caption{Multi-label Classification results using 6 emotion categories}
  \centering
  \begin{tabular}{| c | c  c  c | c  c  c |}
    \hline
       {} &
      \multicolumn{3}{|c|}{F-F1 score} & \multicolumn{3}{|c|}{V-F1 score}  \\ [1ex]
    \hline 
        { Model} & NA & Persian & Comb. & NA & Persian & Comb. 
        \\ [1ex]  \hline
    CC - XGB & \textbf{0.42} & 0.28 & 0.33 & \textbf{0.40} & 0.33 & 0.36
    \\ [1ex] \hline
    MLKNN & \textbf{0.42} & 0.31 & 0.34 & \textbf{0.42} & 0.40 & 0.39
    \\ [1ex] \hline
  \end{tabular}
  
  \label{tab:1}
\end{table}

\begin{table}[t]
\caption{Multi-label classification using 10 emoji categories}
  \centering
  \begin{tabular}{| c | c  c  c | c  c  c |}
    \hline
       {} &
      \multicolumn{3}{|c|}{F-F1 score} & \multicolumn{3}{|c|}{V-F1 score}  \\ [1ex]
    \hline 
        { Model} & NA & Persian & Comb. & NA & Persian & Comb. 
        \\ [1ex]  \hline
    CC - XGB & \textbf{0.24} & 0.22 & 0.20 & 0.28 & \textbf{0.31} & 0.29
    \\ [1ex] \hline
    MLKNN & \textbf{0.27} & 0.22 & 0.25 & 0.28 & 0.27 & \textbf{0.30}
    \\ [1ex] \hline
  \end{tabular}
  
  \label{tab:2}
\end{table}

\begin{table*}[t]
\caption{Baseline of hierarchical classification using 10 emoji labels grouped into 6 labels}
  \centering
  \begin{tabular}{| c | c  c  c | c  c  c |}
    \hline
       {} &
      \multicolumn{3}{|c|}{F-F1 score} & \multicolumn{3}{|c|}{V-F1 score}  \\ [1ex]
    \hline 
        { Model} & NA & Persian & Combined & NA & Persian & Combined 
        \\ [1ex]  \hline
    CC - XGBoost & \textbf{0.34} & 0.28 & 0.21 & 0.32 & \textbf{0.36} & 0.28
    \\ [1ex] \hline
    MLKNN & \textbf{0.38} & 0.30 & 0.28 & \textbf{0.39} & 0.32 & 0.34
    \\ [1ex] \hline
  \end{tabular}
  
  \label{tab:3}
\end{table*}

\subsection{Multi-label Classification}
Our dataset analysis showed multiple labels for many videos - especially emoji labels. Work by Du et al. \cite{DuE1454} also showed that emotions can be mixtures, therefore, we used multi-label, multi-class classifiers for our baseline. Traditional classifiers output a single label for each sample. In order to use them for multi-label classification, we can use either of the following approaches: 1) Adapt existing algorithms to output multiple labels, 2) Transform the problem into another form, e.g. performing binary classification for each label. In the latter case, we can use several classifiers that may or may not be independent from each other. For our baselines, we used Multi-Label K-Nearest Neighbors (MLKNN) \cite{zhang-mlknn} for the adaptive approach and Classifier Chains (CC) \cite{classifier_chains} for the latter. 

\subsection{Classifier Chains}
For a given set of labels $L$ the CC model learns $|L|$ classifiers in which all classifiers are linked in a chain through 22 features. The dataset is transformed in $|L|$ data sets where instances of $j$-th data set has the form $((x_i, l_1, ..., l_{j-1}),l_j), l_j \in {0,1}$.

The advantage of classifier chains method is that it is capable of taking correlations between labels into account while maintaining acceptable computational complexity since the output of previous classifiers is fed into the next ones as additional features. 

\subsection{Training and Testing}
Before the classification step, we separated 25\% of the videos for the test phase. We performed 5-fold cross-validation on the rest to obtain the best order of classifiers for the chain and for the parameter $k$ in MLKNN. In the classifier chains, we used the XGBoost \cite{xgboost} classifier, since it showed promising results on similar datasets~\cite{elderreact}. We input all permutations of the order of labels in the classifier chain and selected the order that yielded the best result on the validation set.
The inputs to the model were AUs, head rotation, and gaze angle for each frame, and then the model outputs set of labels for that frame. We report the sample average of F1-score for all of our models since it is widely used in multi-label classification models.
We computed these metrics in two ways: 1) for each frame, and 2) assigning the label to the whole video by taking the majority of predicted labels on the frames.
The results on emotion labels are presented in Table \ref{tab:1} where CC=Classifier Chains, F-F1 score = Frame-level F1-score, V-F1 score = Video-level F1-score and Comb. = Combined. 

We also combined NA and Persian datasets for all classification models to investigate the result of a culturally heterogeneous dataset. The results are reported under the Combined column in Table \ref{tab:1} to Table \ref{tab:3}. 

\subsection{Baseline Classification}
We performed three types of classification as follows:

\subsubsection{Multi-label Classification Using Emotion Words}
First, we used word labels of emotions (annoyed, anger, hatred, furious, contempt, none) to perform classification.

\subsubsection{Multi-label Classification Using Emojis}
In this experiment, we used 10 emoji labels to perform a multi-label classification. The results are in Table \ref{tab:1} to Table \ref{tab:3}. 

\subsubsection{Hierarchical Classification Using Emojis}
To explore the role of non-lexical labels, we conducted a hierarchical classification using emoji labels. First, we fed in our training and test set into the classifiers similar to what we did with emoji labels, and then mapped each predicted emoji label to its emotion class. The training and test sets were the same as our emotion classification experiment. We used Fig. \ref{fig:common-emojis} to create a mapping of each emoji to its corresponding emotion. We mapped \includegraphics[scale=0.18]{images/unamused.png} or \includegraphics[scale=0.18]{images/smirk.png} to contempt, \includegraphics[scale=0.18]{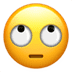} or \includegraphics[scale=0.18]{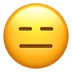} or \includegraphics[scale=0.18]{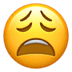} to annoyed, \includegraphics[scale=0.18]{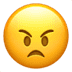} to anger, \includegraphics[scale=0.18]{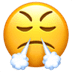} or \includegraphics[scale=0.18]{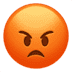} to hatred,  \includegraphics[scale=0.18]{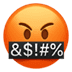} to furious and \includegraphics[scale=0.18]{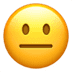} to none.

%
%


 



\subsection{Results}
Overall, NA scores were better than Persian. Video-level scores did not follow a pattern; in some cases they were more than frame-level scores and in some cases less than frame-level. This change in results from frame-level to video-level is highly dependent on test videos. If a test video had a short length (e.g. 50 frames) and the model misclassified a large portion of its frames, it affected video-level scores more than frame-level scores since the number of frames is far greater than the number of videos (2000 frames vs 15 videos in test set). In emotion classification where word labels are used, the difference between NA and Persian is more than emoji and hierarchical classification. In hierarchical and emoji label classification, Persian's video-level F1-score is highest of all.  
Combining the two datasets did not yield better results, usually with results between the two scores of Persian and NA, or less than both, except in emoji classification using the MLKNN model.

\section{Discussion and Limitations}
As shown by the result of data collection, there is always some ambiguity between different emotions. We should also note that the object to which people are reacting may affect the way it is expressed. The result of emoji and hierarchical classification suggests that sometimes non-verbal labels may be better alternative for word labels, especially in cross-cultural research where language might be a barrier. This is very useful in crowd-sourcing platforms like AMT, where labels usually are in English, but annotators' first language may not be. In this study, we cannot completely eliminate the effect of the questionnaire's language. Although people who participated in Persian dataset were competent in English, they likely lacked a deep emotional connection with it and their affective processing may have been weaker \cite{emotion-language}. It was also more difficult to collect high-quality, emotionally-rich videos in Persian due to lack of resources. For example, reaction videos akin to YouTube React channel are almost non-existent for Persian, which makes it an obstacle for this type of research on low-resource cultures. Additionally, we had label-imbalance in our dataset that can negatively affect the result. 
Another technical limitation is the features that we extracted through OpenFace \cite{openface}. OpenFace only extracts 17 AUs out of 30 main AUs.
Finally, emotional expressiveness may affect the result of classifications. Research suggests that more conservative cultures do not express negative emotions such as anger blatantly. We found supporting evidence in the radar charts of Table \ref{tab:radar} that activated AUs in Persian have smaller values. This adds to the complexity of predictive models training on more conservative cultures like Persian.


\section{CONCLUSION AND FUTURE WORK}

\subsection{Conclusion}
The main contribution of this paper is collection of a multi-cultural dataset of videos annotated with (a) affect labels under anger category, (b) their associated emojis, (c) social signals for building more robust emotion recognition models for underrepresented cultures. We conducted statistical analyses to find the underlying expressions of each emotion and compare NA and Persian cultures in terms of emotion expression. Moreover, we provided multi-label classification baseline models that demonstrated how emojis can be used instead of or in addition to word-labels. In order to examine the effectiveness of non-verbal labels, we built a similar model, this time with emoji labels. This opens the opportunity to language-agnostic labels, especially in cross-cultural emotion studies.

\subsection{Future Work}
We collected the presented dataset with intention to classify them as a set of features varying over time (i.e. a multi-feature timeseries). However, there is no machine learning model that is adaptable to multi-label classification of multi-feature timeseries. Hence, a great improvement on the computational aspect of this field would be designing multi-label classifiers that are capable of handling 3-dimensional data (time, features and samples). We are also interested to extend the approach presented here to positive affects such as joy and related emotion such as happiness, surprise and cheerful.
Future work could compare the result using English labels fully translated to Persian (and back-translated to ensure accuracy).
Augmenting the dataset with more videos (synthetically or generated) will allow us to use transfer learning and deep neural network algorithms and investigate their effectiveness.
The purpose of this study was not applying state of the art deep learning models for two main reasons. First, they are black boxes and extracted feature embeddings are harder to interpret (e.g. compared to AUs). Secondly, our dataset is much smaller than the aforementioned datasets and even transfer learning may lead to overfitting on training data. Future research may investigate the accuracy of these models.
We also collected videos from Filipino culture, but due to the low number of annotators we omitted them. We look forward to applying the method to Filipino and other cultural datasets. Training the model on one culture and testing it on another culture can also be a matter of investigation.


{\small
\bibliographystyle{ieee}
\bibliography{egbib}
}

\end{document}